GeoShapley: A Game Theory Approach to Measuring Spatial Effects in Machine Learning Models

Ziqi Li (Ziqi.Li@fsu.edu)

Department of Geography, Florida State University, Tallahassee, United States

**Abstract**

This paper introduces GeoShapley, a game theory approach to measuring spatial effects in machine learning models. GeoShapley extends the Nobel Prize-winning Shapley value framework in game theory by conceptualizing location as a player in a model prediction game, which enables the quantification of the importance of location and the synergies between location and other features in a model. GeoShapley is a model-agnostic approach and can be applied to statistical or black-box machine learning models in various structures. The interpretation of GeoShapley is directly linked with spatially varying coefficient models for explaining spatial effects and additive models for explaining non-spatial effects. Using simulated data, GeoShapley values are validated against known data-generating processes and are used for cross-comparison of seven statistical and machine learning models. An empirical example of house price modeling is used to illustrate GeoShapley's utility and interpretation with real world data. The method is available as an open-source Python package named *geoshapley*.

## 1 Introduction

Machine learning and AI have been increasingly used to model geospatial phenomena with promising performance across various domains. Its strength lies in the capacity to account for complex structures from vast and heterogenous datasets in a scalable, flexible, and accurate manner. Machine learning becomes the preferred choice when the primary goal is in achieving "predictive accuracy" in applications such as land cover classification (Camps-Valls et al., 2013; Zhang et al., 2019), weather forecast (Espeholt et al., 2022; Bi et al., 2023), traffic prediction (Derrow-Pinion et al., 2021), and object detection (Li and Hsu, 2020; Xie et al., 2020). However, from a scientific discovery perspective, geographers are arguably more interested in seeking explanations, aiming to understand the relationships and processes underlying geospatial phenomena. To this end, the black-box nature of machine learning models begins to diminish their utility and overshadow their advantages compared to more interpretable statistical alternatives.

Recent advances in the field of eXplainable AI (XAI) provide a solution to explain black-box machine learning. XAI encompasses a set of processes and methods that enable human users to comprehend and trust the results generated by machine learning algorithms (Das and Rad, 2020). Its objectives are to improve understanding of the underlying decision processes, to enhance credibility and confidence in the model parameters and outcomes, to acknowledge limits and uncertainties, and to inform future model development (Gunning et al., 2019; Murdoch et al., 2019). While the term "XAI" is relatively recent (coined by the US Defense Advanced Research Projects Agency (DARPA) in 2017, Gunning and Aha, 2019), the concept of interpretable machine learning is not novel. Interpretation techniques like permutation feature importance and partial dependence plots have seen extensive application. Permutation feature importance quantifies a feature's contribution to the overall model's predictive accuracy. Removing or altering an important feature is expected to reduce model accuracy, whereas removing an irrelevant feature should have no impact (Breiman, 2001). The partial dependence plot illustrates the marginal effect that one or two features have on the predicted outcome of a machine learning model (Friedman, 2001). It can reveal whether the relationship between the outcome and a feature is linear, non-linear, monotonic, or more intricate. However, most of these interpretation methods are global, meaning they offer only an average explanation of model behavior. In the case of permutation feature importance, a global ranking may not be informative to individuals, as feature contributions may vary across different observations. For instance, an individual applying for a credit card would prefer personalized explanations for the application denial rather than generic statements. This is particularly crucial in cases where compliance with legal requirements is essential, such as the GDPR's right to explanation, which grants individuals the right to request explanations for automated decisions (Goodman & Flaxman, 2017).

In this regard, much of the recent development in XAI focuses on local interpretation as opposed to global, aiming to explain model behaviors for each individual observation. Several Local XAI methods attracted the most attention. The first is LIME (Locally Interpretable Model-agnostic Explanations), and its principle is to consider every complex model as locally linear (Ribeiro et al., 2016). At each observation, by using a simple local surrogate model (e.g., linear regression) with perturbed data drawn from a local neighborhood, a relationship can be approximated locally. However, one of the biggest challenges in LIME is the lack of guidance on how to define a local proximity which the results may be sensitive to (Molnar, 2022). The second development is SHAP (SHapley Additive exPlanations) (Lundberg and Lee, 2017), which unifies several XAI methods (including LIME) into the same Shapley value framework from game theory (Shapley, 1953). Shapley fairly allocates the contribution of individual players in a coalition game with mathematically proven theory and properties. In the machine learning context, we can consider features (i.e., covariates or predictors) as players; the model is a game, and the prediction is the outcome. In this regard, Shapley values provide an additive breakdown of individual prediction to marginal contributions associated with each feature (Štrumbelj and Kononenko, 2014). An expanded discussion on Shapley will be provided in Section 2. Lundberg and Lee (2017) also developed a Shapley-compliant kernel to LIME, so the two methods converge to the same solution. SHAP is employed in industry-leading AI platforms such as Google Vertex and Amazon SageMaker and has been used in various fields. SHAP has also attracted attention in explaining models with geospatial data, including applications in house price modeling (Chen et al., 2020), urban climate (Yuan et al., 2022), travel behavior (Kashifi et al., 2022; Li, 2023), accident analysis (Parsa et al., 2020), disaster susceptibility (Abdollahi and Pradhan, 2023; Zhang et al., 2023), heat-related mortality (Kim and Kim, 2022), among others.

While these examples demonstrate the utility of local interpretation methods in geospatial related applications, it is important to note that most of them primarily focus on explaining non-spatial effects, such as global non-linearity or variable interaction. As a result, the explanations are not explicitly spatial. The shared interest in the concept of 'local' between XAI and geography provides a means to quantify and visualize spatial effects that are intrinsically linked to location (Li, 2022). Spatial effects are inherently interaction effects, where locations interact with nearby ones, resulting in spatial autocorrelation, and location can interact with other features, resulting in spatially varying relationships. Li (2022) found that SHAP, as a local interpretation method, can explain such interactive effects from a machine learning model. By using simulated data with known generating processes that embed true spatial effects, the study revealed that the SHAP-explained XGBoost model can approximate spatial effects in a manner remarkably similar to established spatial statistical methods, specifically the Spatial Lag Model (Anselin, 1988) and Multi-scale Geographically Weighted Regression (Fotheringham et al., 2017). This suggests that XAI can act as a bridge, connecting and converging spatial statistical models with machine learning. It's worth noting that in scenarios where complex spatial and non-spatial effects coexist and are not well-understood, machine learning models may emerge as the preferred choice over traditional spatial statistical approaches. This is due to the challenges inherent in the latter, including assumptions of linearity, model specification and selection, and computational overhead. This discovery presents an opportunity to leverage the advantages of machine learning and XAI to explore more intricate spatial and non-spatial effects. However, there are two major challenges in using the off-the-shelf SHAP method for geospatial data. First, since interaction effects are the key to spatial effects, SHAP only supports computing interaction values for tree-based methods (e.g., decision trees, random forest, gradient boosting trees). This poses a limitation in applying this method to other architectures such as neural networks, which shows promising result for modelling geospatial phenomena recently (Derrow-Pinion et al., 2019; Zhu et al., 2021; Yao and Huang, 2023). Second, location features, such as pairs of coordinates or more complicated location embeddings, should be considered a single joint player in Shapley value calculations. Treating them as separate features would violate the theoretical properties of Shapley value as noted by Harris et al. (2022).

Consequently, this paper will address the aforementioned two challenges by developing GeoShapley, which considers both joint features and interaction effects. GeoShapley can measure spatial and non-spatial effects in a model-agnostic manner that can be applied to explain statistical and machine learning models. This paper will formulate GeoShapley, develop efficient estimation algorithms, and create an open-source Python package. The paper is organized as follows: Section 2 introduces the basics of the Shapley value. Section 3 describes the principles and estimation methods of GeoShapley. Section 4 presents simulation examples to validate GeoShapley against ground truth and to demonstrate its utility across a wide range of statistical and machine learning methods. In Section 5, an empirical example of house price modeling for properties in Greater Seattle, USA, is presented. Section 6 discusses the opportunities and challenges that GeoShapley and XAI bring to geographers. The paper concludes in Section 7.

## 2 Shapley value basics

Shapley values, originate from the field of coalitional game theory, were named in honor of Nobel Prize Laureate Lloyd Shapley, who introduced the concept in his seminal work Shapley (1953). Note that the spelling of "Shapley" is similar to that of the popular GIS Python package "Shapely", and the two should not be confused. The Shapley value considers how to fairly distribute contribution among players participating in a coalition game, and the Shapley value for a player $j$ (denoted as $\varphi_j$) in a game is given by:

$$\varphi_j = \sum_{S \subseteq M \setminus \{j\}} \frac{s!\,(p - s - 1)!}{p!} \big(f(S \cup \{j\}) - f(S)\big) \qquad (1)$$

where $p$ is the total number of players, $M \setminus \{j\}$ is a set of all possible combinations of players excluding $j$, $S$ is a player set in $M \setminus \{j\}$ with a size of $s$, $f(S)$ is the outcome of $S$, and $f(S \cup \{j\})$ is the outcome with players in $S$ plus player $i$. The interpretation of Equation (1) is that the Shapley value of a player is the weighted average of its marginal contribution to the game outcomes, taken over all possible combinations of players. To better illustrate the concept of Shapley value, we can consider a following example. There are three players: A, B, C participate in games and their outcomes are listed in Table 1. The total number of possible combinations is $2^p = 8$.

Table 1. A simple example of Shapley value with three players

| Game | Player set $S$ | Outcome $f(S)$ |
|---|---|---|
| 1 | None | 0 |
| 2 | A | 5 |

| 3 | B | 10 |
| 4 | C | 100 |
| 5 | A, B | 5 |
| 6 | A, C | 120 |
| 7 | B, C | 140 |
| 8 | A, B, C | 150 |

Following Equation (1), the Shapley value for Player A is calculated as follows:

$$\varphi_A = \frac{f(A) - f(None)}{3} + \frac{f(A,B) - f(B)}{6} + \frac{f(A,C) - f(C)}{6} + \frac{f(A,B,C) - f(B,C)}{3}$$

$$= \frac{5}{3} + \frac{-5}{6} + \frac{20}{6} + \frac{10}{3} = 7.5$$

(2)

Reiterate the same calculation, Shapley values for player B and C can be obtained respectively as $\varphi_B = 20$ and $\varphi_C = 122.5$. We can claim that Player C has the strongest contribution to this game.

The Shapley value possesses several useful properties, including efficiency, symmetry, null player, and additivity (Shapley, 1953). Efficiency describes that the sum of individual player contributions equals the outcome when all players participate. In the above example, this is represented as $f(A,B,C) = \varphi_A + \varphi_B + \varphi_C = 150$. Symmetry means that two players will receive the same Shapley value if they have the same marginal contribution in all possible coalitions. Null player asserts that a feature's Shapley value is zero when its marginal contribution to all possible coalitions is zero. Additivity refers to the case when two types of games are combined, the Shapley value of the combined game is the sum of the Shapley values of the individual game. Shapley (1953) formally demonstrated that a solution is both fair and unique under the fulfillment of all four properties.

The concept of Shapley value has been applied to other problems. For example, Lipovetsky and Conklin (2001) extended the Shapley value to linear regression and proposed the Shapley regression value, which determines the contribution from covariates (players) to the model's coefficient of determination ($R^2$ value, outcome). Separate linear regression models are estimated with combinations of covariates, yielding different $R^2$ values. Then, Shapley values can be calculated following Equation (1). This method can fairly evaluate covariate importance in the model under multicollinearity. Furthermore, Štrumbelj and Kononenko (2014) proposed using the Shapley value to explain individual predictions in a prediction model. Under the efficiency property of Shapley values, each single prediction $\hat{y}_i$ can be expressed by:

$$\hat{y}_i = \varphi_0 + \sum_{j=1}^{p} \varphi_{ji} \qquad (3)$$

where $\varphi_{ji}$ is the Shapley value for feature $j$ of observation $i$, and $\varphi_0$ is a base value to adjust for the difference between Shapley values and the prediction value. This expression has a strong connection with additive models (e.g., a linear regression) in which each $\varphi_{ji}$ represents how each feature $j$ marginally contributes to the prediction $\hat{y}_i$. To further illustrate this connection, consider a simple linear system $y = 3 + 2X_1 + X_2$, where $X_1$ and $X_2$ are randomly drawn from a uniform distribution $U(0, 4)$. Figure 1 shows the feature values and their associated Shapley values, where the slopes of the lines represent the coefficients of the linear system. This conveniently links Shapley values to the additive model framework that we are familiar with and provides an intuitive local interpretation for any prediction models.

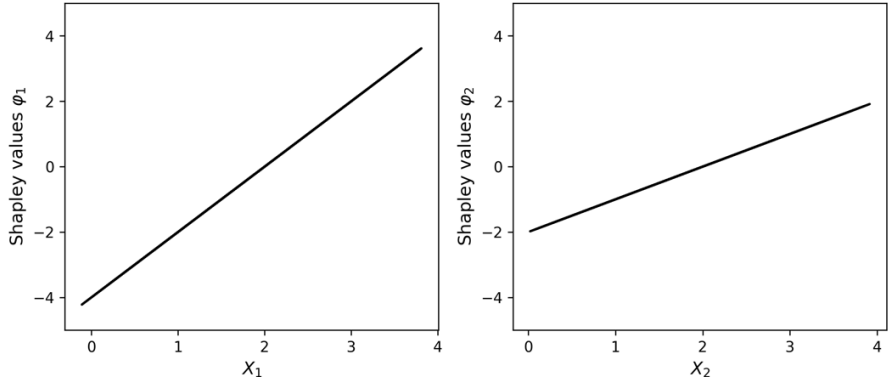

Figure 1. Illustration of the Shapley value and feature value under a linear regression system.

However, the major challenge in adopting the Shapley value empirically, as indicated in Equation 1, is the computational complexity which is exponential in relation to the number of features due to the combinatorial nature of the algorithm. For instance, the calculation becomes intractable when the number of features in a model exceeds 15 ($2^{15} = 32{,}768$ times the prediction function needs to be evaluated). Štrumbelj and Kononenko (2014) developed a Monte Carlo-based approximation approach for estimating the Shapley value to avoid calculating all possible combinations. Furthermore, Lundberg and Lee (2017) proposed a model-agnostic approach named Kernel SHAP, which unifies another popular local interpretation method LIME with Shapley using a Shapley-weighted kernel and can be formulated by solving a weighted least squares regression problem:

$$\boldsymbol{\varphi}_i = \left(\boldsymbol{Z}^T \boldsymbol{W} \boldsymbol{Z}\right)^{-1} \boldsymbol{Z}^T \boldsymbol{W} \boldsymbol{v} \qquad (4)$$

where $\boldsymbol{\varphi_i}$ is a row vector with a size of *1* by *p+1* containing the Shapley values for all *p* features plus a constant for observation *i*; $\boldsymbol{Z}$ is a matrix with a size of $2^p$ by *p+1*, where each row represents a combination of features and is a binary vector with a value of 1 indicating a feature is present and 0 indicating the feature is not. The last column of $\boldsymbol{Z}$ is a constant of 1 used as an intercept when solving the equation which gives us $\boldsymbol{\varphi_0}$. $\boldsymbol{v}$ is the model prediction when using the binary $\boldsymbol{Z}$ matrix to mask the original feature matrix $\boldsymbol{X}$. $\boldsymbol{W}$ is a Shapley weight matrix with a size of $2^p$ by *p+1*. Each row of W, $\boldsymbol{w_s}$ for a feature set *S*, it is given by:

$$w_s = \frac{p-1}{\binom{s}{p} * s * (p-s)} \tag{5}$$

where *p* is the total number of features, *s* is the number of features present in *S*, $\binom{s}{p}$ represents the combination possibilities of choosing *s* out of *p* features. The intuition is that smaller and larger combinations can better isolate individual contributions and should receive higher weights in determining the Shapley value. This is also proportional to the weights in the classic Shapley equation. Lundberg and Lee (2017) also showed that Kernel SHAP has better sample and computational efficiency than the classical Shapley equation and the approach from Štrumbelj and Kononenko (2014). Covert and Lee (2011) further demonstrated that Kernel SHAP is a consistent Shapley value estimator with negligible bias.

Kernel SHAP requires a background dataset. When calculating $\boldsymbol{v}$ with the $\boldsymbol{Z}$ masked feature matrix $\boldsymbol{X}$ as input, it is essential to determine the values that should represent missing features in the pre-trained model, as most models do not explicitly handle missing features without retraining. Therefore, the background dataset is used to approximate the absence of features by replacing the masked-out features with values from this dataset. Options for the background dataset include the entire data, a subset of randomly selected samples, selected representative samples, or a single reference value. The Shapley values computed are then averaged over the background dataset. It should be noted that a larger background dataset reduces the variance in Shapley value estimates but demands more computation. Lundberg and Lee (2017) recommend using the entire dataset for small datasets, and for larger datasets, a single reference point such as the mean or median is preferable, or one may summarize the data with a clustering method such as k-means.

Other Shapley estimators that are worth mentioning include Tree SHAP, which offers the advantage of being significantly more computationally efficient than Kernel SHAP (Lundberg et al., 2021). However, it is only applicable to tree-based models, such as random forests and gradient-boosting trees. In addition to its speed, Tree SHAP can also estimate the Shapley interaction effect, formulated as follows:

$$\phi(i,j) = \sum_{S \subseteq M \backslash\backslash\{i,j\}} \frac{s!\,(p-s-2)!}{2(p-1)!} \Delta_{\{i,j\}}$$

$$\Delta_{\{i,j\}} = f(S \cup \{i,j\}) - f(S \cup \{i\}) - f(S \cup \{j\}) + f(S)$$

<div align="right">(6)</div>

The logic of Equation 6 is similar to Equation 1, but at every feature combination, the interaction effect is isolated from players, $i$ and $j$, by removing their individual contributions, as indicated in $\Delta_{\{i,j\}}$. This interaction effect is particularly useful in giving us an opportunity to measure the interaction between location features and other features and will manifest a location specific effect, as demonstrated in Li (2022). However, Tree SHAP only works for tree-based models. Implementations of various Shapley value estimators are available in open-source Python (shap) and R libraries (shapper and fastshap) and it has been included in popular machine learning packages such as scikit-learn and XGBoost. However, currently, there is no existing model-agnostic approach and tool to estimate Shapley interaction effects. This gap will be addressed in this paper and will be one of its contributions.

## 3 GeoShapley

### 3.1 Principles

Extending from the Shapley value framework, the fundamental concept of GeoShapley, denoted as $\phi$, involves considering location features as a single joint player in the model explanation stage. The concept is that joint players participate in the game together, rather than individually which can be used to combine the location features. Here, location features refer to model inputs that describe the location of a geo-referenced observation, which are often used to account for spatial effects such as spatial autocorrelation and spatial heterogeneity. The simplest case is to use the 2-dimensional coordinates, such as latitude and longitude, as the input in the model. More sophisticated feature transformation methods derived from coordinates have also been found in the literature, such as the use of a combination of Moran eigenvectors (Liu et al., 2017), interactive basis expansions of coordinates (Peaz et al., 2019), Euclidean distance fields (Behrens et al., 2018; Hengl et al., 2018), and space2vec location encoding (Mat et al., 2022). In these cases, coordinates are used to generate high-dimensional features that allow downstream machine learning models to draw decision boundaries more effectively to better characterize the space and achieve higher accuracy. However, high-dimensional features, while useful in the modeling stage, often lack real-world meanings and are not easily interpretable. For instance, instead of understanding the isolated effect of an individual Moran eigenvector or a single high-dimensional feature embedding in the model, it is more relevant to consider their collective contribution to a particular location, analogous to our interpretation of the location effect. Therefore, special consideration is needed when applying the Shapley value framework to location features.

The original Shapley value considers only individual players, and Harris et al. (2021) extended it to accommodate joint players, termed the joint Shapley value, which also satisfies the properties of Shapley values. Following the joint Shapley value, we can calculate $\boldsymbol{\phi_{GEO}}$ as:

$$\boldsymbol{\phi_{GEO}} = \sum_{S \subseteq M \backslash\backslash \{GEO\}} \frac{s!\,(p - s - g)!}{(p - g + 1)!} \big(f(S \cup \{GEO\}) - f(S)\big) \tag{7}$$

where GEO is a set of location features with a size of *g*. If *g=1*, Equation 7 reduces to the individual player situation as in the classic Shapley. *g=2* if including the geographical coordinates (u, v) as the features in the model. This formula is also extendable when *g>2*. The resulting interpretation of $\boldsymbol{\phi_{GEO}}$ is that it measures the marginal contribution to the model prediction from all location features while holding other features constant. This explanation is analogous to a spatial fixed effect or a local intercept in a GWR model, and it is recently termed as an *intrinsic contextual effect* (Fotheringham et al., 2023; Fotheringham and Li ,2023).

Non-spatial effects can be globally linear or non-linear and are location invariant. For a feature $X_j$, its GeoShapley value is formulated as:

$$\boldsymbol{\phi_j} = \sum_{S \subseteq M \backslash\backslash \{j\}} \frac{s!\,(p - s - g)!}{(p - g + 1)!} \big(f(S \cup \{j\}) - f(S)\big) \tag{8}$$

This interpretation is the same as that of the classic Shapley value, which represents the marginal contribution of feature $X_j$ to the model prediction.

Location and feature interaction effect can be calculated based on Shapley Interaction value which is defined as follows:

$$\boldsymbol{\phi_{(GEO,j)}} = \sum_{S \subseteq M \backslash\backslash \{GEO,j\}} \frac{s!\,(p - s - g - 1)!}{(p - g + 1)!} \Delta_{\{GEO,j\}} \tag{9}$$

$$\Delta_{\{Geo,j\}} = f(S \cup \{GEO,j\}) - f(S \cup \{GEO\}) - f(S \cup \{j\}) + f(S)$$

$\boldsymbol{\phi_{(GEO,j)}}$ measures the interaction effect between location and feature *j*, and its interpretation is directly related to a spatially varying coefficient (SVC) models such as GWR and Eigenvector Spatial Filtering. For instance, if GeoShapley is used to explain a GWR model, then $\boldsymbol{\phi_{(GEO,j)}}$ would correspond to the spatially varying component in the model specification, and $(\boldsymbol{\phi(j)} + \boldsymbol{\phi_{(GEO,j)}})/(\boldsymbol{X} - \boldsymbol{E(X_j)})$ gives GWR-type location specific coefficients $\boldsymbol{\beta_j}$ for feature $\boldsymbol{X_j}$. This will be demonstrated in the simulation example discussed later. The formulated GeoShapley retains all the properties of Shapley values by directly applying the joint Shapley and Shapley interaction frameworks.

GeoShapley values are estimated by extending the Kernel SHAP method and are also solved using the weighted least squares framework. In Equation (4), we need to change the binary matrix $\boldsymbol{Z}$ from a size of $2^p$ by $p+1$ to $2^{p-g+1}$ by $2p-2$. The reason is as follows: due to the joint nature of location features, they will be added to the model jointly. This way, we have a smaller combination space than the original one. The exponent represents the size of individual features, which goes from $2^p$ to $2^{p-g+1}$. The column dimension of $\boldsymbol{Z}$ is expanded to include possible interaction effects. Note that two features can only possibly interact if they are both present in the feature set. To better illustrate this, consider that we have $p=4$ features: 2 non-location features, X1 and X2, and 2 location features (e.g., latitude and longitude), where $g=2$, collectively forming the GEO in a prediction model. So, the matrix $\boldsymbol{Z}$ in Equation (4) will have a shape of 7 by 6, as shown in Table 2, where each row represents a possible feature combination, and each column indicates whether a certain feature/combination is present and possible.

Table 2. An example showing the binary Z matrix.

| Feature set | X1 | X2 | X1, GEO | X2, GEO | GEO | Intercept |
|---|---|---|---|---|---|---|
| None | 0 | 0 | 0 | 0 | 0 | 1 |
| X1 | 1 | 0 | 0 | 0 | 0 | 1 |
| X2 | 0 | 1 | 0 | 0 | 0 | 1 |
| GEO | 0 | 0 | 0 | 0 | 1 | 1 |
| X1, GEO | 1 | 0 | 1 | 0 | 1 | 1 |
| X2, GEO | 0 | 1 | 0 | 1 | 1 | 1 |
| X1, X2 | 1 | 1 | 0 | 0 | 0 | 1 |
| X1, X2, GEO | 1 | 1 | 1 | 1 | 1 | 1 |

By following the same estimation procedure as Kernel SHAP, GeoShapley values can be calculated for each individual observation and then averaged over the background dataset. The final output of GeoShapley has four components that add up to the model prediction:

$$\boldsymbol{\hat{y}} = \boldsymbol{\phi_0} + \boldsymbol{\phi_{GEO}} + \sum_{j=1}^{p} \boldsymbol{\phi_j} + \sum_{j=1}^{p} \boldsymbol{\phi_{(GEO,j)}} \qquad (10)$$

where 1) $\phi_0$ is a constant base value. This is the average prediction value given the background data and serves as the global intercept; 2) $\boldsymbol{\phi_{GEO}}$ is a vector with size $n$ measuring the intrinsic location effect in the model; 3) $\boldsymbol{\phi_j}$ is a vector with size $n$ for each non-location feature $j$ giving location-invariant effect to the model; and 4) $\boldsymbol{\phi_{(GEO,j)}}$ is a vector with size $n$ for each non-location feature $j$ giving the spatially varying interaction effect to the model. If no spatial effects are in the model, then the term $\boldsymbol{\phi_{GEO}}$ and $\boldsymbol{\phi_{(GEO,j)}}$ will be zero, and Equation (10) will reduce to Equation (3) proposed by Štrumbelj and Kononenko (2014).

## 3.2 Python package

GeoShapley is accessible as an open-source python package and is distributed in PyPI (The Python Package Index) that can be installed using command `pip install geoshapley`. The package is hosted at [https://github.com/Ziqi-Li/geoshapley](https://github.com/Ziqi-Li/geoshapley). The main function of the package is to compute GeoShapley values for a pre-trained predictive model. The package can be seamlessly integrated with popular machine learning libraries including *scikit-learn*, *xgboost* and *tabnet*. It also offers a variety of visualization tools for visualizing both spatial and non-spatial effects. Tutorials and examples are also included in the code repository. The package will be further maintained and updated as new technical developments of GeoShapley become available. Additionally, data and code to reproduce all the results presented in this paper are also included in the package homepage.

## 4 Simulated examples
## 4.1 Simulation design

This section validates GeoShapley values and their interpretations and provides an example of applying GeoShapley to explain a true model and machine learning models using simulated data. Murdoch et al. (2019) suggested that the evaluation of an explanation method should include simulations to benchmark it against a true model with known Data Generating Processes (DGP). This method allows for the verification of models and explanations can accurately recover the true underlying processes. Following this recommendation, we use a simulation setup that involves a combination of three spatially varying processes, one linear process, and one nonlinear process. This represents a complex scenario encompassing both spatial and non-spatial effects, where spatial processes are location-specific and linear and nonlinear processes are global and location-invariant. The simulation setup largely draws from the work of Harris et al. (2010) and Fotheringham et al. (2017). The approach to model validation and the DGP are shown in Figure 2, with the specific formulas provided as follows:

$$f_0(u, v) = \frac{6}{12^4}\left[12.5^2 - \left(12.5 - \frac{u}{2}\right)^2\right]\left[12.5^2 - \left(12.5 - \frac{v}{2}\right)^2\right]$$

$$f_1(X_1, u, v) = \beta_1(u, v) \circ X_1 \quad where \; \beta_1(u, v) = \frac{2}{49(u + v)}$$

$$f_2(X_2, u, v) = \beta_2(u, v) \circ X_2 \quad where \; \beta_2(u, v) = \frac{2}{49((49 - u) + v)}$$

$$f_3(X_3) = 2X_3$$

$$f_4(X_4) = X_4^2$$

(11)

where $u$ and $v$, ranging from 0 to 49, represent the coordinates of a given location on a 50 by 50 grid, comprising a total of 2,500 geo-referenced observations. $X_1, X_2, X_3, X_4$ are feature inputs. The function $f_0(u, v)$ depends solely on location and is a function of coordinates $u$ and $v$, serving as the intrinsic location effect. Functions $f_1(X_1, u, v)$ and $f_2(X_2, u, v)$ are functions of location and features $X_1$ and $X_2$, respectively. They are denoted as an element-wise multiplication ($\circ$) between spatially varying coefficients $\beta_1(u, v)$ and $\beta_2(u, v)$ and feature value $X_1$ and $X_2$, which manifest as spatially varying effects. Given the challenge of visualizing four-dimensional partial dependence functions (with space as two dimensions, feature as the third, and partial dependence as the fourth), we instead visualize the 2D spatially varying coefficients $\beta_1(u, v)$ and $\beta_2(u, v)$, which is analogous to the interpretation of coefficients in a GWR model. $f_3(X_3)$ and $f_4(X_4)$ represent functions of a linear process (with a slope of 2) and a quadratic U-shaped non-linear effect, respectively.

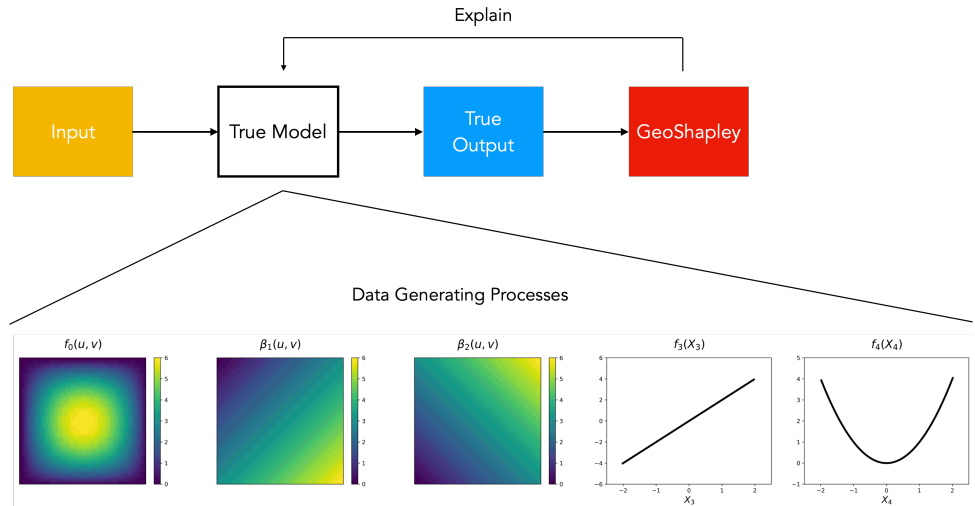

Figure 2. An illustration of the model validation approach and the designed true data generating processes.

## 4.2 True model validation

Given the DGP, we then have a true model with known behaviors. With any input $\boldsymbol{X}$, it will generate a true output $\boldsymbol{y}$. In this example, $\boldsymbol{X}$ are randomly drawn from a uniform distribution of $U(-2, 2)$ as the input to generate a true output $\boldsymbol{y}$ as

$$\boldsymbol{y} = \boldsymbol{f_0}(\boldsymbol{u}, \boldsymbol{v}) + \boldsymbol{f_1}(\boldsymbol{X_1}, \boldsymbol{u}, \boldsymbol{v}) + \boldsymbol{f_2}(\boldsymbol{X_2}, \boldsymbol{u}, \boldsymbol{v}) + \boldsymbol{f_3}(\boldsymbol{X_3}) + \boldsymbol{f_4}(\boldsymbol{X_4}) \qquad (12)$$

GeoShapley values are then computed to explain this true model. We expect the explanations to exactly match the true processes, thereby achieving 100% explanation accuracy. In Figure 3, a summary plot displays the GeoShapley results, ranking the contributions of different features to the true model from top to bottom. The x-axis represents the GeoShapley values, while the feature values are denoted by color. For each feature row, a single dot represents the feature value of each data point (denoted by its color) and its corresponding GeoShapley value (denoted by its horizontal position). Dots accumulate along each feature row to indicate the density distribution. '$GEO$' represents the aggregated contribution of location features. Contributions from non-location features are indicated by their respective names (e.g., X1), and the contribution of the interaction between a feature and location is denoted with a cross (e.g., X1 × GEO). In Figure 3, the GeoShapley values for $\boldsymbol{X_1}$ and $\boldsymbol{X_2}$ are very close in magnitude, suggesting they contribute similarly to the model. This similarity occurs because the data for both are drawn from the same distribution, and the designed location-invariant effect is the same, i.e., $E(\boldsymbol{\beta_1}) = E(\boldsymbol{\beta_2}) = 3$. The GeoShapley value of $\boldsymbol{X_3}$ is approximately two-thirds of that of $\boldsymbol{X_1}$ and $\boldsymbol{X_2}$ because the designed effect $\beta_3 = 2 = \frac{2}{3}E(\boldsymbol{\beta_1}) = E(\boldsymbol{\beta_2})$. Location is the fourth most important feature, as indicated by the GeoShapley values of GEO. $\boldsymbol{X_4}$ has a positive-only contribution since the quadratic function maps all values positively. Regarding interaction effects, the interactions between $\boldsymbol{X_1}$ and location (X1 × GEO), and $\boldsymbol{X_2}$ and location (X2 × GEO), have similar magnitude because the designed spatial patterns in $\boldsymbol{\beta_1}$ and $\boldsymbol{\beta_2}$ are identical, except for the orientation. Interaction effects for $\boldsymbol{X_3}$ and $\boldsymbol{X_4}$ are not found. The summary plot is consistent with our expectations.

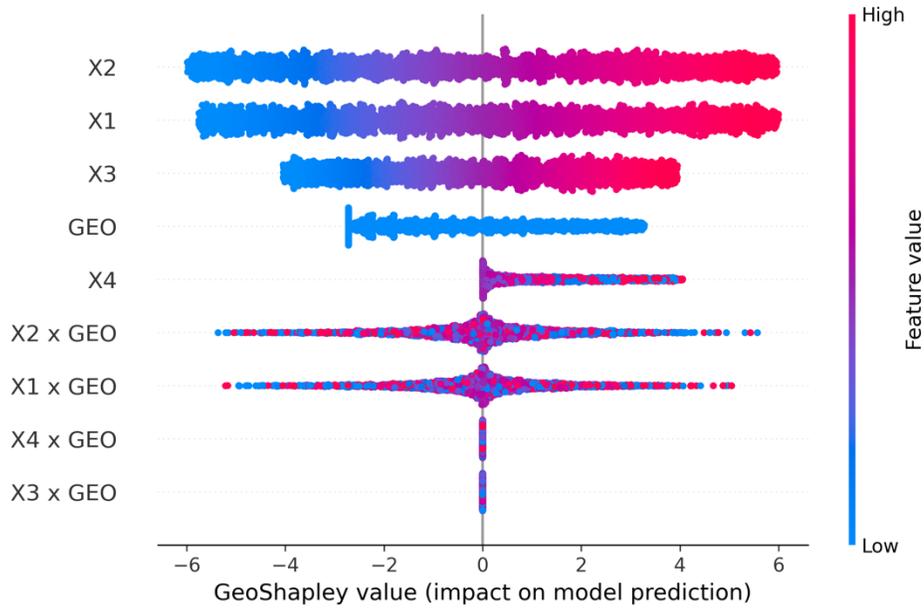

Figure 3. A summary plot showing the feature contribution ranking and distribution of GeoShapley values.

Further calculations can be performed to align the raw GeoShapley values with the true DGP for easier comparison. The intrinsic location effect can be calculated as $\phi_0 + \phi_{GEO}$, which corresponds to $\boldsymbol{f_0(u, v)}$. Spatially varying coefficients can be determined as $\widehat{\boldsymbol{\beta}}_j = \frac{\phi_j + \phi_{(GEO,j)}}{X - E(X_j)}$. No additional calculation is needed for visualizing the non-spatial functions. The explanations resulting from these calculations are displayed in Figure 4, which perfectly matches the original DGP[2]. This demonstrates that when GeoShapley is applied to a true model, it can achieve 100% explanation accuracy.

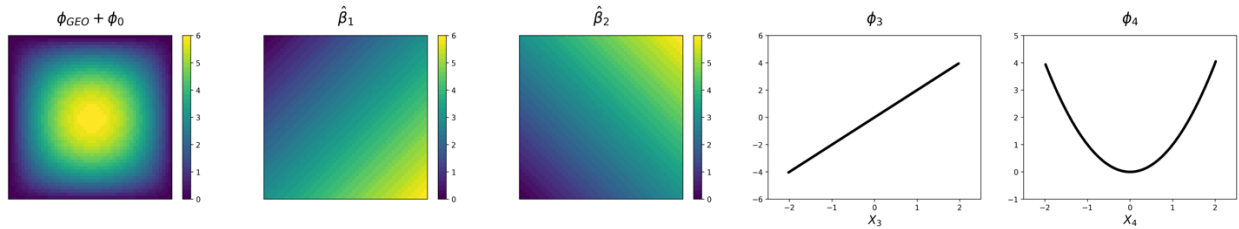

Figure 4. Spatial and non-spatial processes measured by GeoShapley in the true model.

### 4.3 GeoShapley applied to models

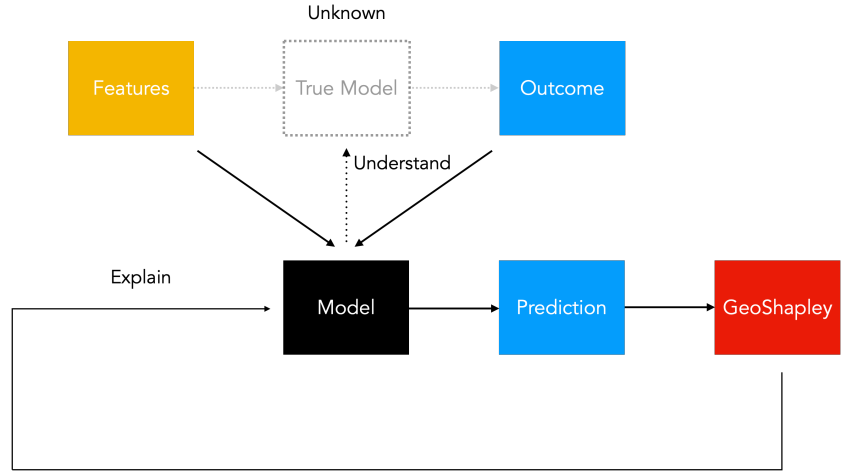

Figure 5. An illustration of model explanation approach to understand true model behavior.

Empirically, as illustrated in Figure 5, a true model is often unknown to us. Instead, we rely on the available data, such as features $\boldsymbol{X}$ and outcome $\boldsymbol{y}$, to fit a model, generate predictions, and use explanation method to explain the model, thereby facilitating our understanding of the underlying processes of the unknown true model. To demonstrate this framework and the utility of GeoShapley, we apply GeoShapley to seven popular statistical and machine learning methods. This evaluation employs a simulated dataset from Equation (12), with a standard normal error added to $\boldsymbol{y}$. The models being compared encompass a wide array of model structures, including:

- Linear Regression (LR)
- Multi-scale Geographically Weighted Regression (MGWR)
- Gaussian Process (GP)
- Random Forest (RF)
- Extreme Gradient Boosting (XGBoost)
- Support Vector Machine (SVM)
- Neural Network (NN)

Due to its model-agnostic nature, GeoShapley conceptually can be applied to any supervised statistical and machine learning models, provided that the original data, predictions, and prediction functions are available. In this comparison, all machine learning models are trained using *scikit-learn* (Pedregosa, 2011), with hyperparameters optimized by the *Hyperopt* Bayesian optimizer (Bergstra, 2013) within the *Hyperopt-sklearn* library. Seventy-five percent of the data is randomly chosen for training, while the remaining twenty-five percent is set aside for testing. For all models, tuning involves 200 evaluations of hyperparameter combinations using the default search space in *Hyperopt-sklearn*. MGWR is estimated with the *mgwr* Python package (Oshan et al., 2019). The goal here is not necessarily to train the best model for the data but rather to evaluate the utility of GeoShapley across empirical models with varying structures. The out-of-sample $R^2$ value, which serves as an indicator of model accuracy, is reported for each model in Table 3. Since MGWR currently lacks an implementation for predicting out-of-sample data, the model $R^2$ is instead reported. The true $R^2$ value is computed based on the added error and the distribution of the ***y***.

Table 3. Model accuracy for the simulated data

|  | Out-of-sample $R^2$ |
| --- | --- |
| LR | 0.757 |
| RF | 0.906 |
| MGWR | 0.943 |
| XGBoost | 0.964 |
| SVM | 0.969 |
| GP | 0.971 |
| NN | 0.971 |
| True $R^2$ | 0.975 |

From Table 3, we observe that linear regression performs the worst, which is anticipated due to its inability to account for spatial and non-linear effects. MGWR shows good predictive performance with an $R^2$ of 0.943, attributable to its capability to capture spatially varying effects. Among the machine learning models, Random Forest exhibits the lowest performance with an $R^2$ of 0.906, while the other models demonstrate comparable performances with $R^2$ values between 0.96 and 0.97, closely approximating the theoretical true $R^2$. The accuracy measures indicate that ML models are very effective in capturing the spatial and non-spatial effects designed in this simulation.

GeoShapley values are calculated for all models, providing explanations for each marginal effect. As with the true model validation, spatial effects are visualized as spatially varying coefficients, making these explanations comparable across models as shown in Figure 6. The first row of Figure 6 represents the true processes. The first three columns display the spatially varying coefficients derived from GeoShapley, while the final two columns show the raw GeoShapley values alongside their corresponding feature values. As expected, the linear regression model only captures constant global linear relationships. The Random Forest model struggles to estimate all effects, showing a noticeable level of noise and inaccuracy, yet it does capture some spatial or non-spatial structures. XGBoost, another tree-based method, outperforms Random Forest by accurately modeling both spatial trends and non-spatial effects. Surprisingly, SVM, GP, and NN perform exceptionally well, nearly perfectly capturing both spatial and non-spatial effects possibly due to their superior handling of continuous decision boundaries compared to XGBoost. The final row shows the coefficients from MGWR capturing the spatial effects and the global linear effect, but not the non-linear U-shaped effect. And it also seems that the inaccuracy in estimating the non-linear effect compromises the estimation of the spatially varying effects. With GeoShapley, this paper is the first to explain spatial effects from machine learning models using a model-agnostic approach, enabling the comparison of these estimated effects across different model structures, which was previously impossible.

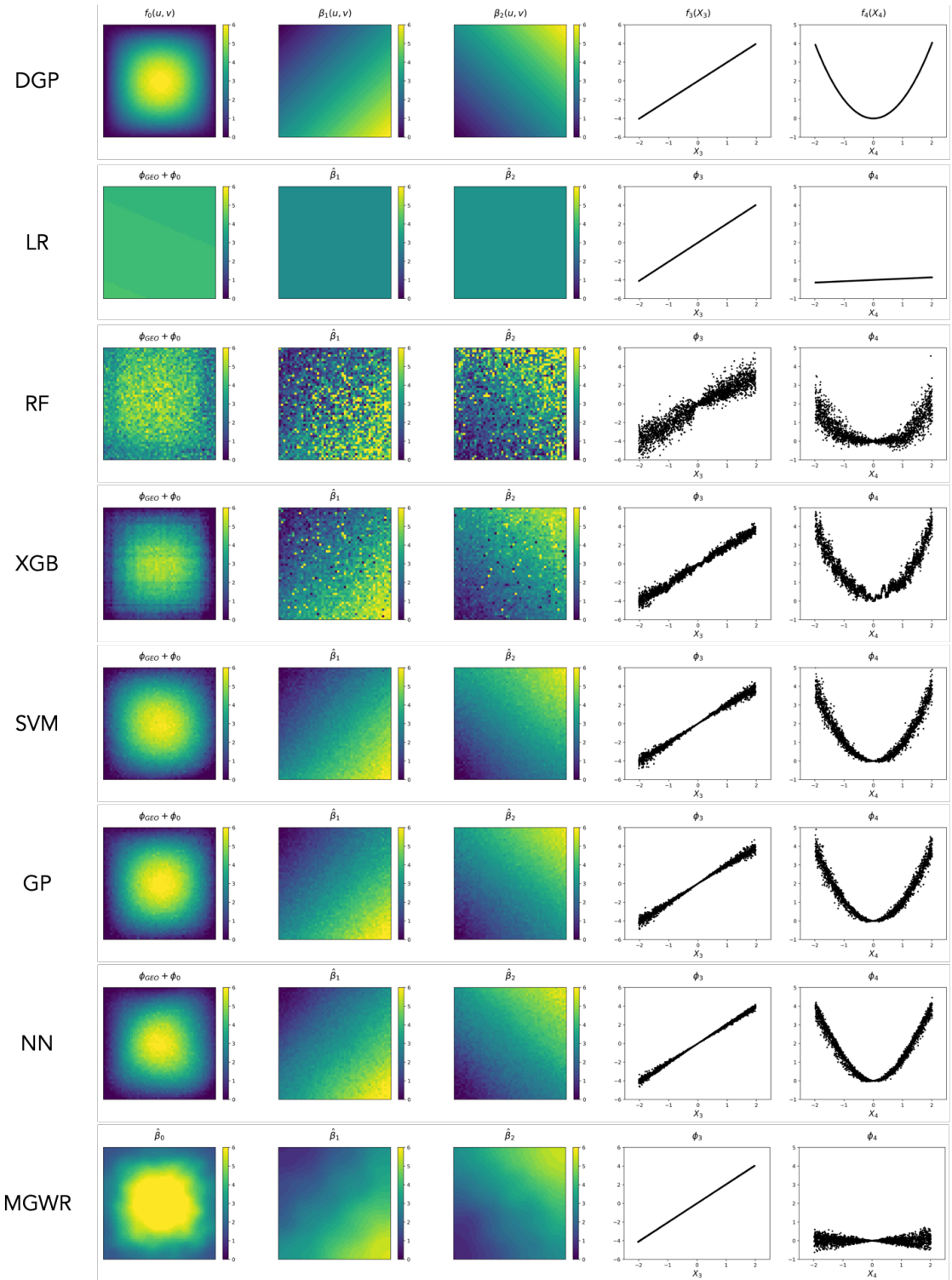

Figure 6. A cross-comparison of explanations obtained from GeoShapley for various models.

## 5 An empirical example of house price modeling

To demonstrate the utility of GeoShapley empirically, we use a case study of housing price modeling in the Greater Seattle area, King County, Washington, USA. This case is selected due to the widely recognized importance of geographic location on home values. This data also has been widely used to demonstrate methodological development in geography (e.g., in Sachdeva et al., 2022 and Anselin and Amaral, 2023). By applying predictive machine learning models along with GeoShapley, we can explain both the effect of location and other features influencing house prices. The dataset, named "Home Sales," is obtained from GeoDa Lab (https://geodacenter.github.io/data-and-lab/) and comprises 21,613 property sales with 21 housing attributes from 2014 to 2015. The data has been cleaned to exclude records with over 10 bedrooms or lots over 20,000 square feet. UTM coordinates are calculated and appended based on the latitude and longitude of each property, and certain properties in extremely rural areas are omitted. The refined dataset includes 16,581 properties. The final model has eight non-location features and two location features. The dependent variable is the logarithm (base 10) of the house sale price. Table 4 provides the descriptions of dependent variable and the features, and Figure 7 shows inter-correlations among the features. Notably, the number of bedrooms, square footage of the living area, and the property grade exhibit moderate to strong bivariate correlations (> 0.6), yet other features show a low level of correlation.

Table 4. Data used in the model and their descriptions.

| Variable | Description |
|---|---|
| Dependent | |
| log_price | The logarithm (base 10) of the sales price |
| | |
| Features | |
| bathrooms | Number of bathrooms in the property[1] |
| sqft_living | Square footage of living area |
| sqft_lot | Square footage of the lot |
| grade | A rank variable from 1 to 13 indicating the construction quality of the property with 1 being the lowest and 13 the highest.[2] |
| condition | A rank variable from 1 to 5 describing the property condition, with 1 indicates poor and worn out, and 5 indicates very well maintained |

---

[1] Properties may have fractional bathrooms that do not have a full set of fixtures.
[2] More information on the grade and condition definitions can be found at:
https://info.kingcounty.gov/assessor/esales/Glossary.aspx

| | |
|---|---|
| waterfront | A binary variable describing whether the property has a waterfront |
| view | An index from 0 to 4 describing how good the view of the property is |
| age | The age of the house as of 2015 |
| UTM_X | The UTM X coordinate of the property |
| UTM_Y | The UTM Y coordinate of the property |

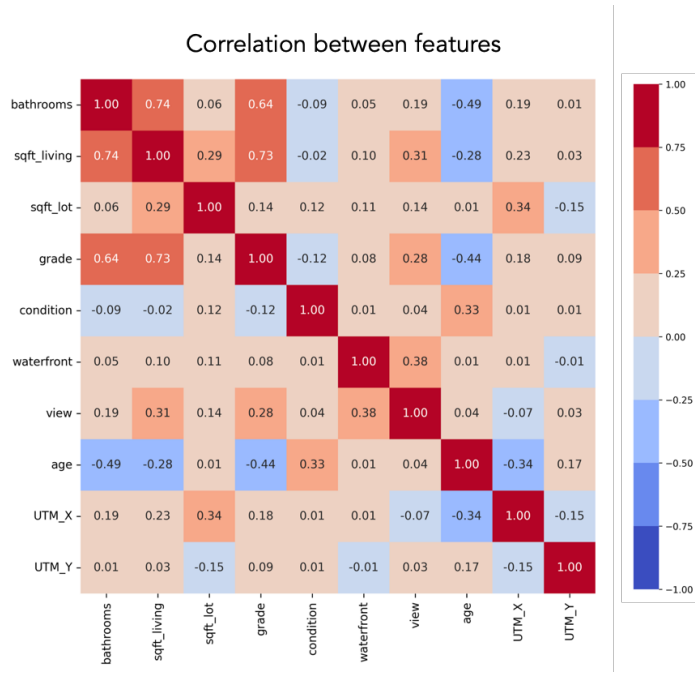

Figure 7. Correlation matrix of the features used in predictive models.

All previously used machine learning models have been trained, and their out-of-sample $R^2$ values are compared in Table 5. In this comparison, XGBoost outperforms all other models. Although the neural network (NN) is the best performer in our simulation example, it is commonly observed that XGBoost yields higher accuracies with real tabular data due to its scalability and flexibility (Shwartz-Ziv and Armon, 2022). The residuals from XGBoost show strong normality and have little-to-no spatial autocorrelation as indicated by a Moran's I value of 0.037. Consequently, we use the trained XGBoost model as our final model to compute GeoShapley values and to explain the effects. Given the size of the dataset, using the entire dataset as the background is computationally feasible yet challenging. Therefore, we use a representative sample of size 50 from k-means clustering. A bootstrap-based approach is adopted to construct confidence intervals around GeoShapley values to capture sampling variability and uncertainties in the model following the work of Efron and Tibshirani (1994). Specifically, bootstrap samples are generated from training data with replacement, and model is re-fitted, and new GeoShapley values are calculated. This process repeats to generate a sampling distribution of GeoShapley values, and a 95% confidence interval can be obtained by finding the 2.5th and 97.5th percentiles of the distribution.

Table 5. Model accuracy for the house price data.

|  | Out-of-sample $R^2$ |
|---|---|
| LR | 0.763 |
| NN | 0.858 |
| GP | 0.869 |
| SVM | 0.878 |
| RF | 0.892 |
| XGBoost | 0.908 |

Since GeoShapley values are in the unit of $\boldsymbol{y}$, which is log-scaled house price, a transformation is applied to convert from the log scale to the percentage change in house price for ease of interpretation. The base value ($\hat{\phi}_0 = 5.63$) indicates that the transformed GeoShapley values are benchmarked against a base property with a price of $10^{5.63}$ or \$430,788. A summary plot ranking feature importance is subsequently presented in Figure 8. Summary statistics of the GeoShapley values are included in Appendix A. The contribution from the location (GEO) emerges as the most important feature affecting a house's price, decreasing it by up to 43% or increasing its value by as much as 123%, depending on the location and holding other factors constant. Housing characteristics, including the square footage of the living area and grade, follow in importance. Features such as a better view and the presence of waterfront can potentially increase a property's value by nearly 50%. There are interactions between location and housing attributes, but they are not as strong as the main non-spatial effects. For example, age shows some degree of interaction with location, suggesting that the impact of age on house price is partly dependent on the property's location.

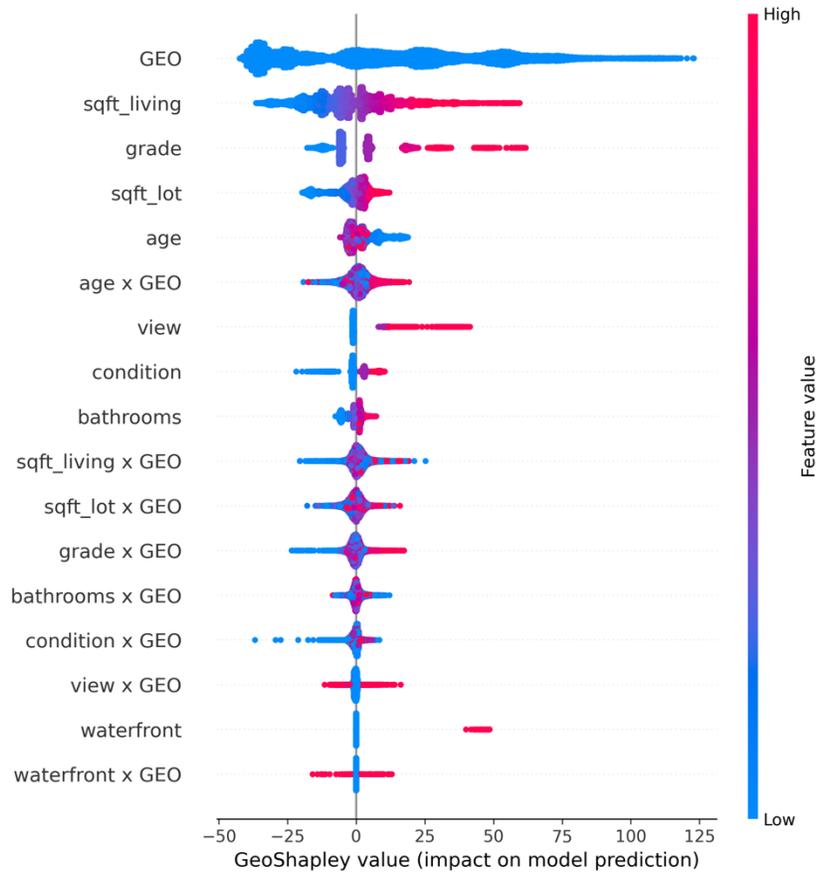

Figure 8. A summary plot showing the feature contribution ranking and distribution of GeoShapley values in the house price model.

The main feature effects, which are independent of location, are shown in Figure 9. Each figure illustrates the marginal contribution to house price in terms of percentage value change when all other features are held constant. These plots help to understand which features have a greater marginal effect on the price of a house and how these relationships are shaped. The bootstrap-based 95% confidence interval is shown as a grey shaded band, and GeoShapley values that have the interval including zero are masked out as grey dots, indicating no statistically significant contribution in the model. Most relationships exhibit a strong degree of non-linearity across the feature's value. For example, as the number of bathrooms increases between 1 and 3, the percentage change in price initially increases, then levels off and becomes non-significant. Increasing both the square footage of the living area and the lot size will significantly increase a property's value, but the impact of the living area is much more substantial than that of the lot size, as indicated by the magnitude of the y-axis. Having a higher grade and better condition will also increase a property's value. The best grade can increase the value by around 60%, while the highest condition can result in a 5-10% increase, *ceteris paribus*. Having a waterfront and the best view can increase the value by roughly 30-40%, respectively. Age generally shows a decreasing trend, with newer houses being valued more than older ones, and age becomes an insignificant factor for most of the houses older than 50.

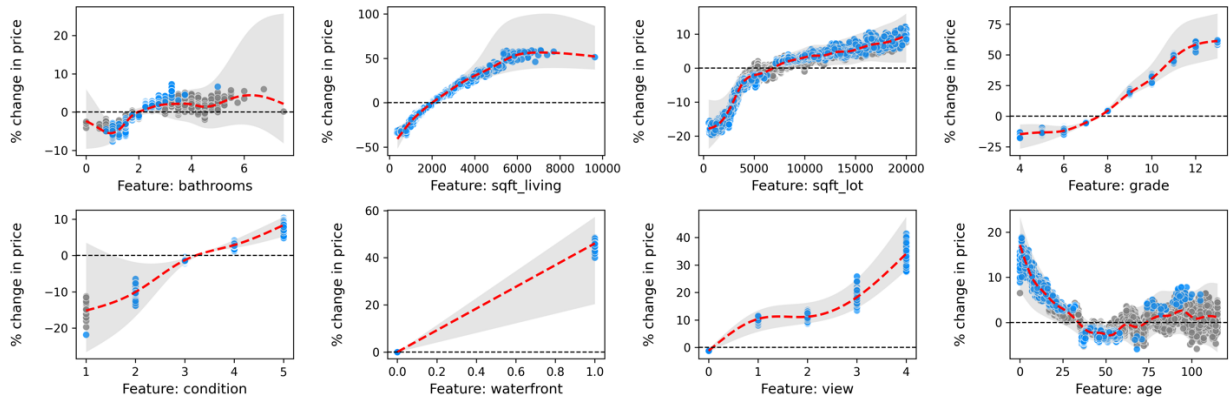

Figure 9. Marginal relationships between housing features and their contribution to the percentage change in house price.

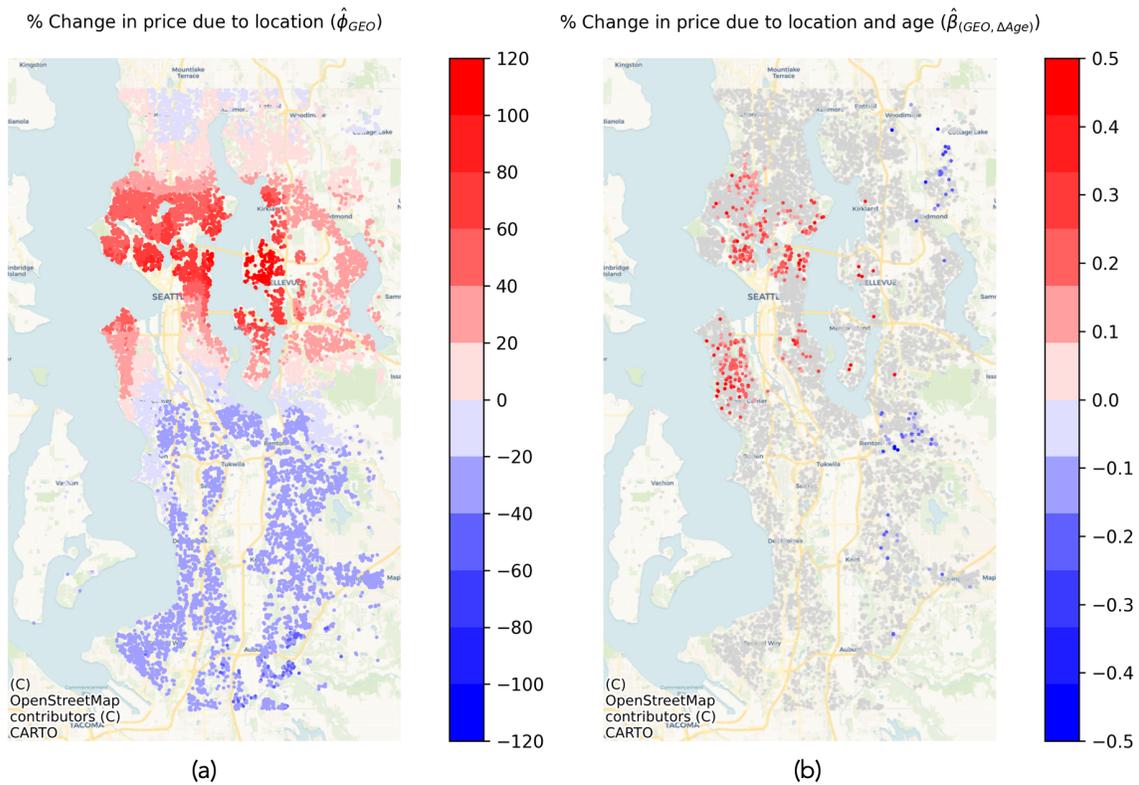

(a)                                                            (b)

Figure 10. Percentage of change in house price due to location (a), and due to the interaction between location and age (b).

The primary spatial effect in this model is the contribution of location to the percentage change in house prices, as measured by the GeoShapley value $\widehat{\phi}_{GEO}$. This contribution ranges from approximately -43% to +123% relative to the base value and is the strongest factor influencing house prices. Figure 10a displays the geographic distribution of this effect. GeoShapley values for which their 95% bootstrap confidence intervals include zero are masked out in grey. The location is expected to act as a proxy for the multifaceted socio-physical context in which people live. It is evident that locations in central and northern Seattle, as well as the Bellevue regions around Lake Washington, substantially and positively influence property prices. These areas are desirable due to their proximity to attractions, businesses, transportation hubs and various physical and urban amenities. Specifically, major employers such as Amazon have been shown to drive up house prices in surrounding neighborhoods (e.g., Bellevue, South Lake Union, Capitol Hill, Queen Anne) (Dowell, 2019). These places also offer good public transportation services and desired amenities such as parks, shopping centers, restaurants, and bars, which enhance the quality of life and contribute to their attractiveness. The University District also holds high location value due to the demand from the University of Washington. In addition, zoning regulations play an important role in housing prices because most residential areas in Seattle, especially north of downtown, are zoned for only single-family housing. This prohibits denser housing and controls supply while demand is increasing in these desirable areas (Glaeser and Gyourko, 2002). Other location-linked factors such as school districts also affect house prices (Franklin and Waddell, 2003), and high location values can be observed within the highly ranked school districts such as Bellevue School District and Mercer Island School District. Conversely, location negatively impacts the price of properties in the southern outskirts of Seattle due to the opposite reasons. In fact, for two properties with identical housing attributes, one in Bellevue would be approximately 3 to 4 times more expensive than a comparable property in the southern rural area of Seattle, owing to the value of the location. The quantified locational value based on GeoShapley also shows very similar spatial pattern when compared to the local intercept estimated from an MGWR model in Sachdeva et al. (2022, Figure 13) and the spatial fixed effect estimated based on a spatially constrained endogenous regimes model in Anselin and Amaral (2023, Figure 2a and Table 7). However, it worth mentioning that the other two approaches are not able to account for the complex nonlinearities arising from the housing attributes compared to machine learning approach.

The second strongest spatial effect is the interaction between location and the age of the property, measured by $\widehat{\phi}_{(GEO,Age)}$. This is further converted to $\widehat{\beta}_{(GEO,\Delta_{Age})}$ and visualized in Figure 10b, which shows how an increase in property age by one year will change the house price in terms of percentage. Similar to the Figure 10a, estimates that are non-significant are masked out in grey. We can see that for most properties, the location-specific age effect is not a significant factor in the model after controlling for its main effect (shown in the last plot of Figure 9). However, a positive association between age and property price is found in neighborhoods such as Queen Anne, Capitol Hill, and West Seattle, indicating that older properties in these areas are valued more than older properties elsewhere. This finding is consistent with Anselin and Amaral (2023), who also found a positive correlation between age and house price in the downtown spatial regime. Additionally, this positive correlation is reported by Rosenberg (2017), who explained that these old homes are in desired neighborhoods and many of them have been recently remodeled that increased their values. Nevertheless, the overall contribution due to interaction between location and age is weak in terms of the magnitude.

## 6 Discussion

From both simulated and empirical examples, it is evident that the GeoShapley value provides a useful framework for explaining spatial and non-spatial effects from machine learning models, which helps to better understand complex processes and relationships underlying geospatial phenomena. The contributions and implications of GeoShapley are summarized below.

Firstly, GeoShapley is model agnostic and works independently of the underlying machine learning model, thanks to the Shapley framework and the Kernel SHAP estimator. This allows for cross-comparison between different methods, ranging from linear regression and spatial models such as MGWR, to more complex machine learning models like XGBoost and Neural Network. The model-agnostic nature also facilitates easier understanding, as it avoids delving into the internal workings of complicated model structures. Additionally, it can be easily integrated into existing machine learning pipelines. For instance, analysts can directly use the *geoshapley* Python package to explain models trained from *scikit-learn*.

Secondly, GeoShapley provides intuitive interpretations of spatial effects, directly linking to the interpretation of additive models and spatially varying coefficient models. This offers a familiar interface for geographers to explore the use of machine learning to understand more complicated spatial and non-linear processes. A notable limitation of many popular spatial statistical models is their assumption of linearity, which can neglect intricate non-spatial effects such as variable interactions and non-linearity. While some of these effects can be addressed through explicit inclusion of interaction terms, non-linear components and data transformations, this still presents a significant model specification and selection challenge, especially when spatial models are often computationally expensive to estimate for large dataset. On the other hand, machine learning models, which are scalable and flexible without strong assumptions about data distribution and functional forms, overcome many of these limitations. Combined with the explanatory capability of GeoShapley, we are better equipped to understand complex relationships in large geospatial datasets that were previously underexplored.

Thirdly, GeoShapley can serve as a diagnostic tool to improve existing machine learning methods for geospatial data. While many recent advancements in GeoAI methodology focus on image-related tasks, relatively few efforts (e.g., Zhu et al., 2021; López and Kholodilin, 2023) have been dedicated to developing methods for tabular data regression, which arguably represents the most common data structure and task that geographers need. With enhanced explanatory capabilities, developers can more precisely identify issues within existing models and propose improvements aimed at enhancing process accuracy rather than just overall model accuracy.

Some limitations and future directions are noteworthy. Firstly, the current implementation of the GeoShapley value lacks formal inferential capabilities. The empirical example demonstrates how to use the bootstrap method to construct confidence intervals, which provides useful information on the uncertainties of the GeoShapley values. This approach inevitably incurs a large computational overhead since the model fitting and explanation need to be repeated many times. However, it is encouraged to do so when the data are moderate or small and if time and hardware allow. Alternatively, GeoShapley can suggest potential functional forms and data transformation that could be integrated into traditional statistical models when formal statistical inference is preferred. Furthermore, Joseph (2022) introduced a parametric inferential framework that uses the Shapley-Taylor decomposition combined with a surrogate parametric regression analysis. Additionally, Covert and Lee (2021) proposed an approximate Kernel SHAP estimator and developed a method to calculate its uncertainties. However, the extension of these approaches to GeoShapley remains an area for further exploration. Secondly, there has been increasing interest in spatial causality to better understand the 'why' question (Akbari et al., 2023; Credit and Lehnert, 2023; Gao et al., 2023; Zhang and Wolf, 2023). Janzing et al. (2020) argued that the Shapley value inherently possesses a causal interpretation when using an interventional algorithm (what Kernel SHAP and GeoShapley use), which examines how a model behaves when the dependence among features is deliberately disrupted. The Shapley value also provides an interface to investigate more complex causal effects within machine learning models and has been shown to be effective in measuring heterogeneous treatment effects with instrumental variables (Syrgkanis et al., 2019). Furthermore, Heskes et al. (2020) introduced causal Shapley values, a model-agnostic approach to measure the direct and indirect effects from each feature to the model's prediction. It is useful to extend and apply these methods to geospatial tasks.

## 7 Conclusion

This paper introduces GeoShapley value as an explanation tool to measure spatial effects in machine learning models. GeoShapley extends from the classic Shapley value framework in game theory by considering location features as a player in a model prediction game. The GeoShapley value is formulated based on joint Shapley and Shapley interaction frameworks and is estimated by Kernel SHAP. GeoShapley is capable of quantifying the intrinsic location effect in the model, the interaction effect between location and other features, and the remaining linear or nonlinear effects. The interpretation of GeoShapley is directly linked with spatially varying coefficient models for explaining spatial effects. The explanation accuracy of GeoShapley is validated against a true model with known data-generating processes, and its utility is demonstrated in both simulated data and real-world data. GeoShapley is available as an open-source Python package *geoshapley*.

**Author Biographies**

**ZIQI Li** is an Assistant Professor in the Department of Geography at Florida State University, Tallahassee, FL 32306. E-mail: Ziqi.Li@fsu.edu. His research interests include spatial statistical modeling and interpretable machine learning.

**List of Figure Captions**

Figure 1. Illustration of the Shapley value and feature value under a linear regression system.

Figure 2. An illustration of the model validation approach and the designed true data generating processes.

Figure 3. A summary plot showing the feature contribution ranking and distribution of GeoShapley values.

Figure 4. Spatial and non-spatial processes measured by GeoShapley in the true model.

Figure 5. An illustration of model explanation approach to understand true model behavior.

Figure 6. A cross-comparison of explanations obtained from GeoShapley for various models.

Figure 7. Correlation matrix of the features used in predictive models.

Figure 8. A summary plot showing the feature contribution ranking and distribution of GeoShapley values in the house price model.

Figure 9. Marginal relationships between housing features and their contribution to the percentage change in house price.

Figure 10. Percentage of change in house price due to location (a), and due to the interaction between location and age (b).

**Appendix A**

Table A1: Summary statistics of GeoShapley values for the Seattle example.

| GeoShapley value | Min. | 25% | 50% | 75% | Max. |
|---|---|---|---|---|---|
| GEO | -43.48 | -24.68 | 0.02 | 33.21 | 123.17 |
| bathrooms | -7.60 | -3.08 | -0.07 | 1.25 | 7.30 |
| sqft_living | -36.45 | -12.65 | -3.61 | 6.84 | 59.53 |
| sqft_lot | -19.63 | -2.63 | 0.57 | 2.65 | 12.13 |
| grade | -17.84 | -5.75 | -5.04 | 4.34 | 61.77 |
| condition | -21.82 | -1.36 | -1.16 | 2.84 | 10.39 |
| waterfront | 0.00 | 0.00 | 0.00 | 0.00 | 48.59 |
| view | -1.46 | -1.26 | -1.21 | -1.13 | 41.44 |
| age | -5.90 | -1.77 | 0.71 | 3.78 | 18.88 |
| GEO, bathrooms | -8.52 | -0.59 | 0.06 | 0.93 | 12.14 |
| GEO, sqft_living | -20.51 | -0.55 | 0.71 | 2.24 | 25.26 |
| GEO, sqft_lot | -17.85 | -1.65 | -0.20 | 1.16 | 15.95 |
| GEO, grade | -23.48 | -1.34 | -0.28 | 0.75 | 17.46 |
| GEO, condition | -36.83 | -0.98 | -0.09 | 0.44 | 8.46 |
| GEO, waterfront | -15.85 | 0.00 | 0.00 | 0.00 | 12.99 |
| GEO, view | -11.48 | -0.59 | -0.25 | 0.08 | 16.24 |
| GEO, age | -19.2 | -1.07 | 0.97 | 3.07 | 19.26 |

**Appendix B**

An experiment based on the validation dataset is conducted to examine the sampling variance of GeoShapley values as a function of background sample size. As can be seen in Figure B1, the sampling variance decreases drastically after sample size of 20 and becomes negligible after 50.

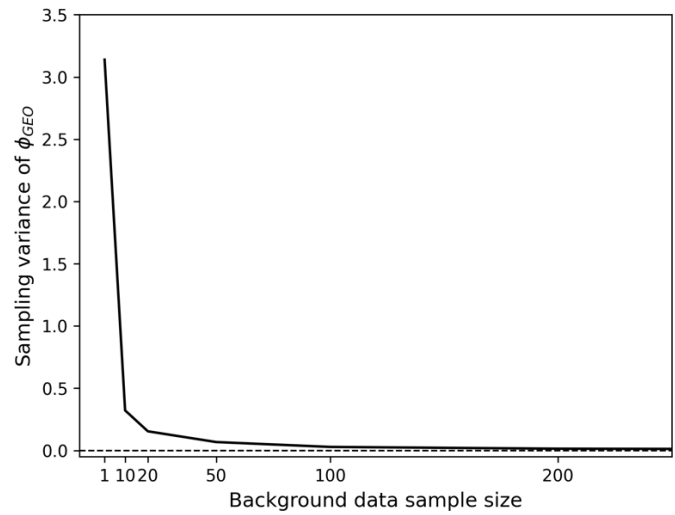

Figure B1. The effect of background sample size on sampling variance of $\boldsymbol{\phi_{GEO}}$